\documentclass[a4paper,fleqn]{cas-sc}

\usepackage[authoryear,longnamesfirst]{natbib}
\usepackage{booktabs}
\usepackage{multirow}
\usepackage{graphicx}  
\usepackage{array}     
\usepackage{subfig}

\usepackage[section]{placeins}
\usepackage{flafter}
\usepackage{float}
\makeatletter
\def\fps@figure{!htbp}
\def\fps@table{!ht}
\makeatother
\def\tsc#1{\csdef{#1}{\textsc{\lowercase{#1}}\xspace}}
\tsc{WGM}
\tsc{QE}

\begin{document}
\let\WriteBookmarks\relax
\raggedbottom

\shorttitle{}    

\shortauthors{Wenjie Ou et~al.}

\title [mode = title]{Logo-LLM: Local and Global Modeling with Large Language Models for Time Series Forecasting}  

\tnotemark[1]
\author[1]{Wenjie Ou}
\cormark[1]
\ead{ouwenjie@stu.scu.edu.cn}
\affiliation[1]{organization={College of Computer Science, Sichuan University},
    city={Chengdu},
    postcode={610065}, 
    country={China}}
    
\author[1]{Zhishuo Zhao}[style=chinese]
\cormark[1]
\ead{zhaozhishuo@stu.scu.edu.cn}

\author[1]{Cheng Chen}[style=chinese]
\ead{2024223045001@stu.scu.edu.cn}

\author[1]{Dongyue Guo}[style=chinese]
\ead{dongyueguo@scu.edu.cn}

\author
[1]
{Yi Lin}
\cormark[2]
\ead{yilin@scu.edu.cn}

\cortext[cor1]{Equal contribution to this work.}
\cortext[cor2]{Corresponding author.}
\tnotemark[1,2]


\begin{abstract}
Time series forecasting is critical across multiple domains, where time series data exhibit both local patterns and global dependencies. While Transformer-based methods effectively capture global dependencies, they often overlook short-term local variations in time series. Recent methods that adapt large language models (LLMs) into time series forecasting inherit this limitation by treating LLMs as black-box encoders,  relying solely on the final-layer output and underutilizing hierarchical representations. To address this limitation, we propose Logo-LLM, a novel LLM-based framework that explicitly extracts and models multi-scale temporal features from different layers of a pre-trained LLM. Through empirical analysis, we show that shallow layers of LLMs capture local dynamics in time series, while deeper layers encode global trends. Moreover, Logo-LLM introduces lightweight Local-Mixer and Global-Mixer modules to align and integrate features with the temporal input across layers. Extensive experiments demonstrate that Logo-LLM achieves superior performance across diverse benchmarks, with strong generalization in few-shot and zero-shot settings while maintaining low computational overhead.
\end{abstract}


\begin{highlights}
\item Reveals layer-wise specialization of LLMs for local and global time patterns.
\item First to exploit multi-layer LLM features to decouple local and global series.
\item Designs Local-Mixer and Global-Mixer to align LLMs with temporal data.
\item Logo-LLM improves long-term forecasting and few-/zero-shot generalization.
\item Achieves strong accuracy–efficiency trade-off with competitive complexity.

\end{highlights}


\maketitle

\section{Introduction}
 Time series forecasting is a critical task across various domains, where temporal data naturally exhibit multi-scale patterns, including short-term local variations and long-range global dependencies. Local features capture transient behaviors within short time windows, while global dependencies reflect correlations across broader temporal spans. Early Transformer-based models (~\cite{liu2022non, zhou2022fedformer, wu2021autoformer,liu2021pyraformer}) have emerged as the dominant approach due to their remarkable capacity to model global dependencies via the attention mechanism. However, these methods often neglect local temporal patterns, limiting their ability to capture fine-grained fluctuations. To address this, ~\cite{Yuqietal2023PatchTST} proposes a patch-based framework that significantly models local variations, emphasizing short-term dynamics in time series data. This perspective has spurred a series of subsequent studies (\cite{chen2024pathformer, zhang2024multi}) aiming to explicitly capture local and global temporal patterns in a multi-scale manner.

\begin{figure}[t]
  \centering
  \includegraphics[width=\columnwidth,height=0.22\textheight,keepaspectratio]{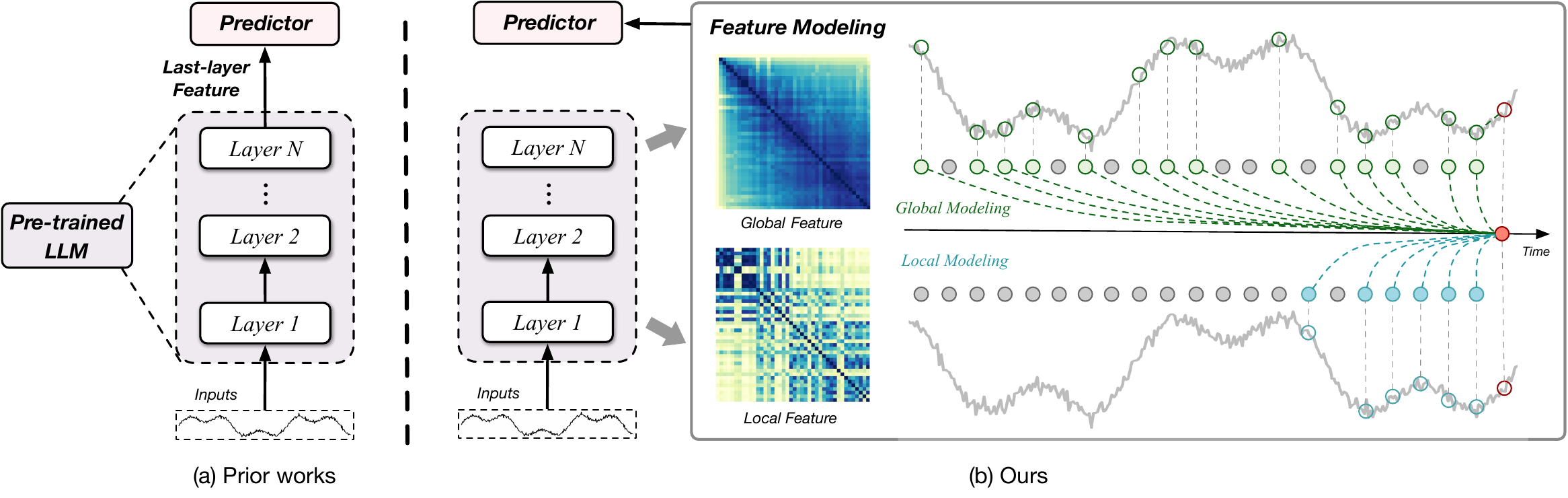}
  \caption{Comparison of LLM usage paradigms. Prior works treat LLMs as black-box encoders and use only the last-layer feature. Our method explicitly extracts features from multiple layers, leveraging shallow-layer features for local modeling and deep-layer features for global modeling, enabling a more fine-grained understanding of temporal dynamics.}
  \label{fig:logo}
\end{figure}

Despite these advancements, recent efforts to adapt large language models (LLMs) for time series (~\cite{liu2025calf, jin2023timellm, zhou2023onefitsall}) appear to inherit the same limitations. While LLMs inherit the global modeling strengths of Transformer, existing approaches typically treat the LLM as a black-box encoder, leveraging only the final-layer output for prediction. This practice underutilizes the rich hierarchical representations distributed across different layers, resulting in a persistent bias towards global patterns and a neglect of local temporal variations. To apply the LLM more effectively, we first analyze the internal representational properties of LLMs in the context of time series. As shown in Figure \ref{fig:logo}, our empirical analysis reveals a clear layer-wise specialization: shallow layers are more sensitive to local dynamics, while deeper layers encode broader temporal dependencies. This insight motivates us to design a new paradigm that explicitly extracts and integrates multi-layer LLM features, leveraging shallow-layer outputs as local representations and deep-layer outputs as global representations.

Inspired by the above insights, we propose a \textbf{Lo}cal and \textbf{g}l\textbf{o}bal modeling \textbf{LLM}-based framework called \textbf{Logo-LLM} that models local and global features through LLM's representations. Logo-LLM employs a pre-trained GPT2 (~\cite{radford2019gpt2}) backbone to extract multi-layer semantic representations, explicitly using the early-layer features to capture local temporal variations and the deep-layer features to encode global dependencies. To effectively align these heterogeneous features with temporal data, we design two specialized modules: the Global-Mixer, which integrates global representations to model long-range dependencies, and the Local-Mixer, which aligns local representations to capture short-term fluctuations. Decoupling local and global modeling enables more precise capture of multi-scale temporal patterns, addressing the limitation of relying solely on final-layer outputs. The experimental results on real-world benchmarks demonstrate that Logo-LLM consistently achieves superior performance in long-term forecasting, with strong generalization in few-shot and zero-shot settings under limited data. Notably, our approach maintains competitive complexity, offering an efficient yet powerful alternative for temporal modeling with LLMs.

Here we summarize our key contributions as follows:
\begin{enumerate}
    \item We propose a Local and global modeling LLM-based framework (Logo-LLM) to achieve superior performance in time series forecasting tasks supported by extensive experiments, including long-term, few-shot, and zero-shot scenarios. To the best of our knowledge, this study is the first to propose leveraging the internal layer-wise semantics of a pre-trained LLM to separately capture local and global variations in time series data.
    \item We design Local-Mixer and Global-Mixer modules to separately align local and global features with the temporal data and capture long- and short-term variations in time series. Decoupling local and global features allows the model to better exploit the hierarchical nature of LLMs rather than rely solely on the final-layer output.
    \item Extensive experiments on multiple real-world benchmark datasets demonstrate that Logo-LLM achieves superior performance in long-term forecasting, few-shot, and zero-shot learning tasks, with favorable generalization ability and low computational complexity.
\end{enumerate}

\section{Related Work}
\subsection{LLMs for Time Series}
With the great popularity of LLMs in the NLP field, the application of LLMs into time series tasks has emerged as an innovative and promising approach. LLM-based methods leverage the extensive world knowledge acquired through pre-training to enhance contextual modeling capabilities in time series analysis. ~\cite{zhou2023onefitsall} demonstrated the potential of fine-tuning and adapting LLMs for time series forecasting. ~\cite{jin2023timellm} reprogrammed time series inputs by text-based prototypes and augmented them by treating the context as a prefix to achieve improved alignment with LLMs. ~\cite{liu2025calf} proposed a framework that incorporates specific modules to align textual data with temporal data. Despite these advances, these methods ignore the intrinsic local features of temporal data, resulting in the full potential of LLMs' general knowledge being underutilized.

\subsection{Local and Global Modeling for Time Series}
Time series data inherently exhibit both local and global variations. Transformer-based architectures (~\cite{liu2022non, zhou2022fedformer, wu2021autoformer, Yuqietal2023PatchTST}) have demonstrated superior ability in capturing global patterns. However, standard Transformers suffer from high computational costs for long sequences and may overlook crucial local details. To address this, ~\cite{micn} adopts a multi-scale isometric convolutional network to separately capture local and global contexts. In the CV and NLP domain, recent studies (~\cite{li2022locality, chen2023building, lee2024causal, liu2024fantastic}) observed that Transformer models naturally learn fine-grained local textures in shallow layers, while deep layers capture high-level semantics and global dependencies. Motivated by this insight, we introduce a hierarchical feature modeling paradigm, progressing from local to global representations, into time series forecasting.

\section{Methodology}
\begin{figure}[htbp]
  \centering
  \includegraphics[width=\columnwidth]{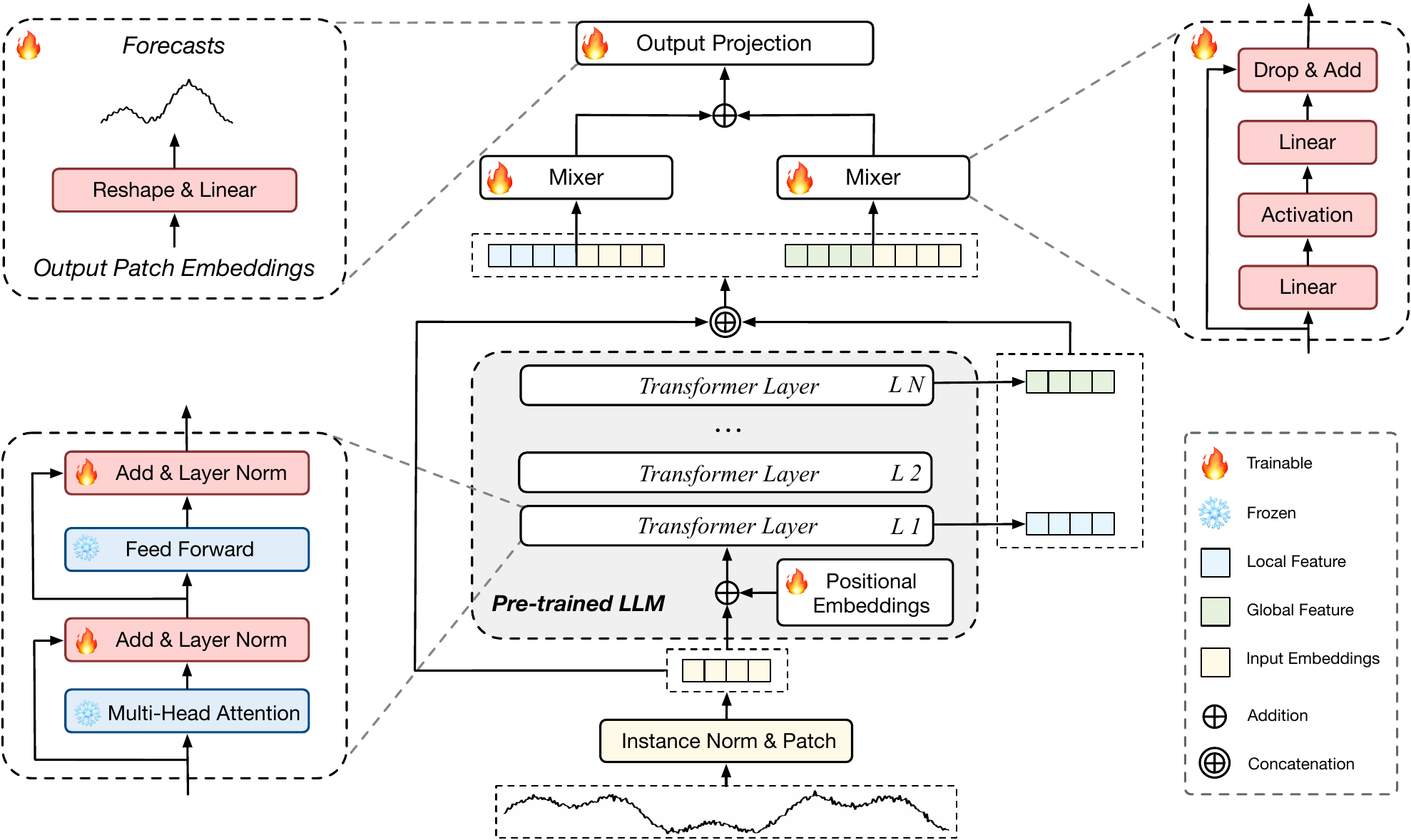}
  \caption{Overview of the proposed Logo-LLM framework. Logo-LLM extracts intermediate representations from multiple layers of a pre-trained LLM to explicitly model local and global temporal patterns. Two specialized Mixer modules are introduced to align these hierarchical features with the temporal input, enabling fine-grained modeling of local and global variations. Most LLM parameters are kept frozen, enabling efficient adaptation under limited supervision.}
  \label{fig:model}
\end{figure}
Our model architecture is illustrated in Figure \ref{fig:model}. Given a time series input $\textbf{X} \in \mathbb{R}^{L\times D}$ of length $L$, our goal is to leverage the multi-layer features generated by a pre-trained LLM. Specifically, the input is first processed through instance normalization and patching, and then passed into the LLM to obtain hierarchical features. Unlike previous works that treat LLMs as black-box encoders using only final-layer features, we decouple temporal modeling by explicitly leveraging shallow-layer and deep-layer outputs. These features are concatenated with the patching input and fed into corresponding Mixer modules, which serve to align the input with the learned temporal representations and enhance the modeling of temporal dynamics. Finally, a fusion step integrates the aligned features by an output projection layer.

\noindent \textbf{Input Transformations.}\hspace{0.2cm}
Given a multivariate time series $\textbf{X} \in \mathbb{R}^{L\times D}$, we split it into $D$ univariate time series $\mathbf{X}_{i} \in \mathbb{R}^{L \times 1}$ along the channel dimension and process them independently, following the Channel-Independence strategy proposed in ~\cite{Zeng2022dlinear}. To ensure a similar distribution, we apply instance normalization (~\cite{kim2021reversible}) to each univariate series. The normalization is defined as:
\begin{align}
\begin{split}
        \mu_i = \frac{1}{L}\sum_{t=1}^{L}\mathbf{X}_{i,t}, \quad \sigma_i = \frac{1}{L}\sum_{t=1}^{L}(\mathbf{X}_{i,t}-\mu_i), \quad \bar{\textbf{X}}_{i, t} = \frac{\textbf{X}_{i, t} - \mu_i}{\sqrt{\sigma_i + \epsilon}},
\end{split}
\end{align}
where $\mu_{i}, \sigma_i$ denote the mean and variance of the $t$-th time step of the $i$-th univariate series and $\epsilon$ is a small constant for numerical stability. Inspired by ~\cite{Yuqietal2023PatchTST}, we further segment each normalized series $\bar{\textbf{X}_i}$ into overlapping patches of length $P$ and stride $S$. The technique can allow the model to see the longer contextual sequence, which can significantly improve prediction performance.
\begin{align}
    \bar{\textbf{X}}_{i,j} = \mathrm{Unfold}(\mathrm{ReplicationPad}(\bar{\textbf{X}}_i), \mathrm{size}=P, \mathrm{step}=S),
\end{align}
where $\bar{\textbf{X}}_{i,j}$ is the $j$-th patch of the $i$-th input series.

\noindent \textbf{Pre-trained LLM Backbone.}\hspace{0.2cm}We design a lightweight input embedding layer to adapt raw temporal data for integration with the pre-trained LLM. The embedding layer linearly projects each patch into the token space, formulated as $\tilde{\textbf{X}}_i=\mathrm{TE}(\bar{\textbf{X}}_{i})$, where $\mathrm{TE}(\cdot)$ denotes a learnable token embedding. To preserve temporal ordering, we also incorporate a positional embedding layer $\mathrm{PE}(\cdot)$. The embedded inputs are then fed into $N$ Transformer blocks of a pre-trained LLM, with only $\mathrm{PE}(\cdot)$ and layer normalization layers fine-tuned to significantly reduce the number of trainable parameters. Notably, self-attention and FFN layers remain untouched to retain the rich knowledge obtained through pre-training. The process is described as:
\begin{align}
\begin{split}
        \bar{\textbf{H}}_{i} &= \tilde{\textbf{X}}_i + \mathrm{PE}(\bar{\textbf{X}}_{i}), \\
    \bar{\textbf{H}}_{i}^{(n)} &= \mathcal{T}^{(n)}(\bar{\textbf{H}}_{i}^{(n-1)}), \quad n \in \{1, \cdot\cdot\cdot, N\},
\end{split}
\end{align}
where $\mathcal{T}^{(n)}, \bar{\textbf{H}}_{i}^{(n)}$ are the $n$-th Transformer block and corresponding hidden state of the $i$-th patch.

\noindent \textbf{Mixer Modules.}\hspace{0.2cm}For each patch, the pre-trained LLM captures temporal dependencies through stacked Transformer layers, progressively encoding local and global patterns. Inspired by hierarchical representation learning in vision models, we leverage intermediate LLM outputs to disentangle temporal features at different levels of abstraction. Specifically, we extract shallow-layer feature $\bar{\textbf{H}}_{i}^{(0)}$ to represent local short-term variations and deep-layer feature $\bar{\textbf{H}}_{i}^{(N)}$ to capture broader temporal trends. This design provides a clearer semantic separation between local and global temporal patterns. As illustrated in Figure \ref{fig:logo}, shallow-layer outputs preserve high-resolution temporal details, while deeper layers abstract long-range patterns, justifying our decoupled design. Moreover, the design improves computational efficiency by avoiding aggregation across multiple layers.

Furthermore, we employ two lightweight MLP-based Mixer modules (Local-Mixer and Global-Mixer) to align these hierarchical features with the input patches. These modules enhance non-linearity and fuse semantic representations from the LLM with temporal sequences, facilitating more accurate forecasting. The fused results are then aggregated through element-wise addition and projected via a linear layer to produce the final predictions.
\begin{align}
\begin{split}
    \mathbf{\bar{H}}_{i,l} &= \tilde{\mathbf{X}}_i + \mathrm{Dropout}\left( \mathbf{W}_l^{(2)} \cdot \phi\left( \mathbf{W}_l^{(1)} \cdot [\tilde{\mathbf{X}}_i \, \Vert \, \bar{\textbf{H}}_{i}^{(0)}] \right) \right),  \\
\mathbf{\bar{H}}_{i,g} &= \tilde{\mathbf{X}}_i + \mathrm{Dropout}\left( \mathbf{W}_g^{(2)} \cdot \phi\left( \mathbf{W}_g^{(1)} \cdot [\tilde{\mathbf{X}}_i \, \Vert \, \bar{\textbf{H}}_{i}^{(N)}] \right) \right),  \\
\mathbf{Y}_i &= \mathbf{W}_{out} \cdot \mathrm{Reshape}\left( \mathbf{\bar{H}}_{i,l} + \mathbf{\bar{H}}_{i,g} \right),
\end{split}
\end{align}
where $[:\, \Vert \,:]$ and $\phi$ denote $\mathrm{Concat}(\cdot)$ and $\mathrm{ReLU(\cdot)}$ functions. $\mathbf{W}_l^{(1)}, \mathbf{W}_l^{(2)}$ and $\mathbf{W}_g^{(1)}, \mathbf{W}_g^{(2)}$ are the MLP weights in the Local-Mixer and Global-Mixer modules, respectively. $\mathbf{W}_{out}$ denotes the weight matrix of output projection layer. $\textbf{Y}_i$ is the final results of the $i$-th patch.

\section{Experimental Results}

\noindent \textbf{Experimental Settings.}\hspace{0.2cm}In various benchmarks, Logo-LLM consistently outperforms state-of-the-art time series forecasting methods, especially in few-shot and zero-shot learning. We compare our approach with a wide range of recent models, especially fine-tuning time series analysis language models, including CALF (~\cite{liu2025calf}), Time-LLM (~\cite{jin2023timellm}), and GPT4TS (~\cite{zhou2023onefitsall}). Meanwhile, we also select several recent competitive models, including iTransformer (~\cite{liu2023itransformer}), PatchTST (~\cite{Yuqietal2023PatchTST}), DLinear (~\cite{Zeng2022dlinear}), TimesNet (~\cite{wu2023timesnet}), and FEDformer (~\cite{zhou2022fedformer}). We use ADAM as the default optimizer and report the mean squared error (MSE) and mean absolute error (MAE) as the evaluation metrics. Specifically, we use GPT2 (~\cite{radford2019gpt2}) as the default backbone to validate the effectiveness of Logo-LLM. We use ADAM as the default optimizer and report the mean squared error (MSE) and mean absolute error (MAE) as the evaluation metrics. For models whose default input length is not 96, we modify only the input length to 96 while keeping other settings unchanged. To ensure a fair comparison, we adopt the official implementations of the backbone models and follow their default configurations.

\noindent \textbf{Datasets.}\hspace{0.2cm}The datasets are characterized as follows. We follow standard protocol of (\cite{Yuqietal2023PatchTST}) and split all datasets into training, validation, and test sets in chronological order by the ratio of 6:2:2 for the ETT dataset and 7:1:2 for the other datasets.
\begin{itemize}
    \item \textbf{ETT}\hspace{0.2cm}contains two sub-datasets: ETT1 and ETT2, collected from two electricity transformers at two stations. Each of them has two versions in different resolutions (15min \& 1h). The ETT dataset contains multiple series of loads and one series of oil temperatures.

\item \textbf{Electricity Consuming Load (ECL)}\hspace{0.2cm}corresponds to the electricity consumption (Kwh) of 321 clients.

\item \textbf{Weather}\hspace{0.2cm}contains 21 meteorological indicators, such as air temperature, humidity, etc, recorded every 10 minutes for the entirety of 2020. 

\item \textbf{Exchange}\hspace{0.2cm}collects the daily exchange rates of 8 countries (Australia, British, Canada, Switzerland, China, Japan, New Zealand, and Singapore) from 1990 to 2016. 

\item \textbf{Traffic} includes data on the occupancy rates of the freeway system, recorded from 862 sensors across the State of California, with the sampling rate of 1 hour.

\item \textbf{National Illness (ILI)}\hspace{0.2cm}corresponds to the weekly recorded influenza-like illness patients from the US Center for Disease Control and Prevention. 
\end{itemize}

\begin{table}[pos=htbp]\scriptsize
\centering
\caption{Long-term time series forecasting. Best results are in \textbf{bold} and second are in \underline{underlined}.}
\label{tab:long-term}
\setlength{\tabcolsep}{2.4pt}
\linespread{1}
\begin{tabular}{@{}cc|cc|cc|cc|cc|cc|cc|cc|cc|cc@{}}
\toprule
\multicolumn{2}{c}{Methods}                                                & \multicolumn{2}{c|}{\makecell{Logo-LLM\\ (Ours)}}                                                 & \multicolumn{2}{c|}{CALF} & \multicolumn{2}{c|}{Time-LLM} & \multicolumn{2}{c|}{GPT4TS} & \multicolumn{2}{c|}{iTransformer} & \multicolumn{2}{c|}{PatchTST}                                                 & \multicolumn{2}{c|}{TimesNet} & \multicolumn{2}{c|}{FEDformer} & \multicolumn{2}{c}{DLinear} \\ \midrule
\multicolumn{2}{c|}{Metric}                                                & MSE                                   & \multicolumn{1}{c|}{MAE}                                   & MSE                                   & \multicolumn{1}{c|}{MAE}                                   & MSE            & \multicolumn{1}{c|}{MAE}   & MSE                                   & \multicolumn{1}{c|}{MAE}         & MSE                                   & \multicolumn{1}{c|}{MAE}                                   & MSE                                   & \multicolumn{1}{c|}{MAE}                                   & MSE         & \multicolumn{1}{c|}{MAE}   & MSE   & \multicolumn{1}{c|}{MAE}   & MSE             & MAE       \\ \midrule
\multicolumn{1}{c|}{}                           & \multicolumn{1}{c|}{96}  & {\color[HTML]{000000} \textbf{0.317}} & \multicolumn{1}{c|}{{\color[HTML]{000000} \textbf{0.343}}} & \underline {0.323}                           & \multicolumn{1}{c|}{\underline {0.350}}                           & 0.359          & \multicolumn{1}{c|}{0.381} & 0.329                                 & \multicolumn{1}{c|}{0.364}       & 0.341                                 & \multicolumn{1}{c|}{0.376}                                 & 0.328                                 & \multicolumn{1}{c|}{0.367}                                 & 0.338       & \multicolumn{1}{c|}{0.375} & 0.379 & \multicolumn{1}{c|}{0.419} & 0.345           & 0.372     \\
\multicolumn{1}{c|}{}                           & \multicolumn{1}{c|}{192} & {\color[HTML]{000000} \textbf{0.368}} & \multicolumn{1}{c|}{{\color[HTML]{000000} \textbf{0.370}}} & \underline {0.374}                           & \multicolumn{1}{c|}{\underline {0.375}}                           & 0.383          & \multicolumn{1}{c|}{0.393} & {\color[HTML]{000000} \textbf{0.368}} & \multicolumn{1}{c|}{0.382}       & 0.382                                 & \multicolumn{1}{c|}{0.395}                                 & {\color[HTML]{000000} \textbf{0.368}} & \multicolumn{1}{c|}{0.385}                                 & \underline {0.374} & \multicolumn{1}{c|}{0.387} & 0.426 & \multicolumn{1}{c|}{0.441} & 0.380           & 0.389     \\
\multicolumn{1}{c|}{}                           & \multicolumn{1}{c|}{336} & {\color[HTML]{000000} \textbf{0.394}} & \multicolumn{1}{c|}{{\color[HTML]{000000} \textbf{0.393}}} & 0.409                                 & \multicolumn{1}{c|}{\underline {0.399}}                           & 0.416          & \multicolumn{1}{c|}{0.414} & 0.400                                 & \multicolumn{1}{c|}{0.403}       & 0.418                                 & \multicolumn{1}{c|}{0.418}                                 & \underline {0.399}                           & \multicolumn{1}{c|}{0.410}                                 & 0.410       & \multicolumn{1}{c|}{0.411} & 0.445 & \multicolumn{1}{c|}{0.459} & 0.413           & 0.413     \\
\multicolumn{1}{c|}{\multirow{-4}{*}{ETTm1}}    & \multicolumn{1}{c|}{720} & {\color[HTML]{000000} \textbf{0.460}} & \multicolumn{1}{c|}{{\color[HTML]{000000} \textbf{0.430}}} & 0.477                                 & \multicolumn{1}{c|}{\underline {0.438}}                           & 0.483          & \multicolumn{1}{c|}{0.449} & 0.460                                 & \multicolumn{1}{c|}{0.439}       & 0.487                                 & \multicolumn{1}{c|}{0.456}                                 & \underline {0.454}                           & \multicolumn{1}{c|}{0.439}                                 & 0.478       & \multicolumn{1}{c|}{0.450} & 0.543 & \multicolumn{1}{c|}{0.490} & 0.474           & 0.453     \\ \midrule
\multicolumn{1}{c|}{}                           & \multicolumn{1}{c|}{96}  & {\color[HTML]{000000} \textbf{0.175}} & \multicolumn{1}{c|}{{\color[HTML]{000000} \textbf{0.253}}} & \underline {0.178}                           & \multicolumn{1}{c|}{\underline {0.256}}                           & 0.193          & \multicolumn{1}{c|}{0.280} & \underline {0.178}                           & \multicolumn{1}{c|}{0.263}       & 0.185                                 & \multicolumn{1}{c|}{0.272}                                 & 0.183                                 & \multicolumn{1}{c|}{0.270}                                 & 0.187       & \multicolumn{1}{c|}{0.267} & 0.203 & \multicolumn{1}{c|}{0.287} & 0.193           & 0.292     \\
\multicolumn{1}{c|}{}                           & \multicolumn{1}{c|}{192} & {\color[HTML]{000000} \textbf{0.243}} & \multicolumn{1}{c|}{{\color[HTML]{000000} \textbf{0.298}}} & \underline {0.245}                           & \multicolumn{1}{c|}{\underline {0.300}}                           & 0.257          & \multicolumn{1}{c|}{0.318} & \underline {0.245}                           & \multicolumn{1}{c|}{0.306}       & 0.253                                 & \multicolumn{1}{c|}{0.313}                                 & 0.255                                 & \multicolumn{1}{c|}{0.314}                                 & 0.533       & \multicolumn{1}{c|}{0.563} & 0.269 & \multicolumn{1}{c|}{0.328} & 0.284           & 0.362     \\
\multicolumn{1}{c|}{}                           & \multicolumn{1}{c|}{336} & {\color[HTML]{000000} \textbf{0.304}} & \multicolumn{1}{c|}{{\color[HTML]{000000} \textbf{0.338}}} & \underline {0.309}                           & \multicolumn{1}{c|}{\underline {0.341}}                           & 0.317          & \multicolumn{1}{c|}{0.353} & \underline {0.309}                           & \multicolumn{1}{c|}{0.347}       & 0.315                                 & \multicolumn{1}{c|}{0.350}                                 & \underline {0.309}                           & \multicolumn{1}{c|}{0.347}                                 & 0.321       & \multicolumn{1}{c|}{0.351} & 0.325 & \multicolumn{1}{c|}{0.366} & 0.369           & 0.427     \\
\multicolumn{1}{c|}{\multirow{-4}{*}{ETTm2}}    & \multicolumn{1}{c|}{720} & \underline {0.408}                           & \multicolumn{1}{c|}{\underline {0.400}}                           & {\color[HTML]{000000} \textbf{0.402}} & \multicolumn{1}{c|}{{\color[HTML]{000000} \textbf{0.395}}} & 0.419          & \multicolumn{1}{c|}{0.411} & 0.409                                 & \multicolumn{1}{c|}{0.408}       & 0.413                                 & \multicolumn{1}{c|}{0.406}                                 & 0.412                                 & \multicolumn{1}{c|}{0.404}                                 & \underline {0.408} & \multicolumn{1}{c|}{0.403} & 0.421 & \multicolumn{1}{c|}{0.415} & 0.554           & 0.522     \\ \midrule
\multicolumn{1}{c|}{}                           & \multicolumn{1}{c|}{96}  & \underline {0.373}                           & \multicolumn{1}{c|}{{\color[HTML]{000000} \textbf{0.388}}} & {\color[HTML]{000000} \textbf{0.369}} & \multicolumn{1}{c|}{\underline {0.389}}                           & 0.398          & \multicolumn{1}{c|}{0.410} & 0.376                                 & \multicolumn{1}{c|}{0.397}       & 0.386                                 & \multicolumn{1}{c|}{0.404}                                 & 0.393                                 & \multicolumn{1}{c|}{0.408}                                 & 0.384       & \multicolumn{1}{c|}{0.402} & 0.376 & \multicolumn{1}{c|}{0.419} & 0.386           & 0.400     \\
\multicolumn{1}{c|}{}                           & \multicolumn{1}{c|}{192} & {\color[HTML]{000000} \textbf{0.413}} & \multicolumn{1}{c|}{{\color[HTML]{000000} \textbf{0.412}}} & \underline {0.427}                           & \multicolumn{1}{c|}{\underline {0.423}}                           & 0.451          & \multicolumn{1}{c|}{0.440} & 0.438                                 & \multicolumn{1}{c|}{0.426}       & 0.441                                 & \multicolumn{1}{c|}{0.436}                                 & 0.445                                 & \multicolumn{1}{c|}{0.434}                                 & 1.008       & \multicolumn{1}{c|}{0.792} & 0.420 & \multicolumn{1}{c|}{0.448} & 0.437           & 0.432     \\
\multicolumn{1}{c|}{}                           & \multicolumn{1}{c|}{336} & {\color[HTML]{000000} \textbf{0.434}} & \multicolumn{1}{c|}{{\color[HTML]{000000} \textbf{0.424}}} & \underline {0.456}                           & \multicolumn{1}{c|}{\underline {0.436}}                           & 0.508          & \multicolumn{1}{c|}{0.471} & 0.479                                 & \multicolumn{1}{c|}{0.446}       & 0.489                                 & \multicolumn{1}{c|}{0.461}                                 & 0.484                                 & \multicolumn{1}{c|}{0.451}                                 & 0.491       & \multicolumn{1}{c|}{0.469} & 0.459 & \multicolumn{1}{c|}{0.465} & 0.481           & 0.459     \\
\multicolumn{1}{c|}{\multirow{-4}{*}{ETTh1}}    & \multicolumn{1}{c|}{720} & {\color[HTML]{000000} \textbf{0.446}} & \multicolumn{1}{c|}{{\color[HTML]{000000} \textbf{0.447}}} & \underline {0.479}                           & \multicolumn{1}{c|}{\underline {0.467}}                           & 0.483          & \multicolumn{1}{c|}{0.478} & 0.495                                 & \multicolumn{1}{c|}{0.476}       & 0.508                                 & \multicolumn{1}{c|}{0.493}                                 & 0.480                                 & \multicolumn{1}{c|}{0.471}                                 & 0.521       & \multicolumn{1}{c|}{0.500} & 0.506 & \multicolumn{1}{c|}{0.507} & 0.519           & 0.516     \\ \midrule
\multicolumn{1}{c|}{}                           & \multicolumn{1}{c|}{96}  & {\color[HTML]{000000} \textbf{0.274}} & \multicolumn{1}{c|}{{\color[HTML]{000000} \textbf{0.330}}} & \underline {0.284}                           & \multicolumn{1}{c|}{\underline {0.336}}                           & 0.295          & \multicolumn{1}{c|}{0.346} & 0.295                                 & \multicolumn{1}{c|}{0.348}       & 0.300                                 & \multicolumn{1}{c|}{0.349}                                 & 0.294                                 & \multicolumn{1}{c|}{0.343}                                 & 0.340       & \multicolumn{1}{c|}{0.374} & 0.358 & \multicolumn{1}{c|}{0.397} & 0.333           & 0.387     \\
\multicolumn{1}{c|}{}                           & \multicolumn{1}{c|}{192} & {\color[HTML]{000000} \textbf{0.344}} & \multicolumn{1}{c|}{{\color[HTML]{000000} \textbf{0.376}}} & \underline {0.353}                           & \multicolumn{1}{c|}{\underline {0.380}}                           & 0.386          & \multicolumn{1}{c|}{0.399} & 0.386                                 & \multicolumn{1}{c|}{0.404}       & 0.379                                 & \multicolumn{1}{c|}{0.398}                                 & 0.377                                 & \multicolumn{1}{c|}{0.393}                                 & 0.402       & \multicolumn{1}{c|}{0.414} & 0.429 & \multicolumn{1}{c|}{0.439} & 0.477           & 0.476     \\
\multicolumn{1}{c|}{}                           & \multicolumn{1}{c|}{336} & \underline {0.369}                           & \multicolumn{1}{c|}{\underline {0.398}}                           & {\color[HTML]{000000} \textbf{0.362}} & \multicolumn{1}{c|}{{\color[HTML]{000000} \textbf{0.394}}} & 0.447          & \multicolumn{1}{c|}{0.443} & 0.421                                 & \multicolumn{1}{c|}{0.435}       & 0.418                                 & \multicolumn{1}{c|}{0.429}                                 & 0.381                                 & \multicolumn{1}{c|}{0.409}                                 & 0.452       & \multicolumn{1}{c|}{0.452} & 0.496 & \multicolumn{1}{c|}{0.487} & 0.594           & 0.541     \\
\multicolumn{1}{c|}{\multirow{-4}{*}{ETTh2}}    & \multicolumn{1}{c|}{720} & {\color[HTML]{000000} \textbf{0.394}} & \multicolumn{1}{c|}{{\color[HTML]{000000} \textbf{0.417}}} & \underline {0.406}                           & \multicolumn{1}{c|}{\underline {0.428}}                           & 0.428          & \multicolumn{1}{c|}{0.444} & 0.422                                 & \multicolumn{1}{c|}{0.445}       & 0.428                                 & \multicolumn{1}{c|}{0.445}                                 & 0.412                                 & \multicolumn{1}{c|}{0.433}                                 & 0.462       & \multicolumn{1}{c|}{0.468} & 0.463 & \multicolumn{1}{c|}{0.474} & 0.831           & 0.657     \\ \midrule
\multicolumn{1}{c|}{}                           & \multicolumn{1}{c|}{24}  & 1.683                                 & \multicolumn{1}{c|}{{\color[HTML]{000000} \textbf{0.800}}} & {\color[HTML]{000000} \underline {1.672}}    & \multicolumn{1}{c|}{\underline {0.841}}                           & \textbf{1.651} & \multicolumn{1}{c|}{0.841} & 1.869                                 & \multicolumn{1}{c|}{\underline {0.823}} & 2.321                                 & \multicolumn{1}{c|}{0.937}                                 & 2.221                                 & \multicolumn{1}{c|}{0.883}                                 & 1.826       & \multicolumn{1}{c|}{0.893} & 2.721 & \multicolumn{1}{c|}{1.133} & 5.060           & 1.709     \\
\multicolumn{1}{c|}{}                           & \multicolumn{1}{c|}{36}  & 1.825                                 & \multicolumn{1}{c|}{{\color[HTML]{000000} \textbf{0.816}}} & {\color[HTML]{000000} \underline {1.725}}    & \multicolumn{1}{c|}{0.872}                                 & \textbf{1.701} & \multicolumn{1}{c|}{0.861} & 1.853                                 & \multicolumn{1}{c|}{\underline {0.854}} & 2.188                                 & \multicolumn{1}{c|}{0.945}                                 & 2.313                                 & \multicolumn{1}{c|}{0.904}                                 & 2.678       & \multicolumn{1}{c|}{0.986} & 2.768 & \multicolumn{1}{c|}{1.118} & 4.413           & 1.549     \\
\multicolumn{1}{c|}{}                           & \multicolumn{1}{c|}{48}  & {\color[HTML]{000000} \textbf{1.777}} & \multicolumn{1}{c|}{{\color[HTML]{000000} \textbf{0.793}}} & 1.937                                 & \multicolumn{1}{c|}{0.937}                                 & 2.153          & \multicolumn{1}{c|}{1.041} & \underline {1.886}                           & \multicolumn{1}{c|}{\underline {0.855}} & 2.231                                 & \multicolumn{1}{c|}{0.956}                                 & 2.048                                 & \multicolumn{1}{c|}{0.886}                                 & 2.584       & \multicolumn{1}{c|}{0.937} & 2.637 & \multicolumn{1}{c|}{1.088} & 4.109           & 1.473     \\
\multicolumn{1}{c|}{\multirow{-4}{*}{ILI}}      & \multicolumn{1}{c|}{60}  & {\color[HTML]{000000} \textbf{1.748}} & \multicolumn{1}{c|}{{\color[HTML]{000000} \textbf{0.807}}} & 2.128                                 & \multicolumn{1}{c|}{0.999}                                 & 2.064          & \multicolumn{1}{c|}{0.953} & \underline {1.877}                           & \multicolumn{1}{c|}{\underline {0.877}} & 2.292                                 & \multicolumn{1}{c|}{0.991}                                 & 2.008                                 & \multicolumn{1}{c|}{0.915}                                 & 1.980       & \multicolumn{1}{c|}{0.894} & 2.696 & \multicolumn{1}{c|}{1.050} & 4.233           & 1.481     \\ \midrule
\multicolumn{1}{c|}{}                           & \multicolumn{1}{c|}{96}  & \underline {0.173}                           & \multicolumn{1}{c|}{{\color[HTML]{000000} \textbf{0.206}}} & {\color[HTML]{000000} \textbf{0.168}} & \multicolumn{1}{c|}{\underline {0.207}}                           & 0.195          & \multicolumn{1}{c|}{0.233} & 0.182                                 & \multicolumn{1}{c|}{0.223}       & 0.174                                 & \multicolumn{1}{c|}{0.214}                                 & 0.177                                 & \multicolumn{1}{c|}{0.218}                                 & 0.172       & \multicolumn{1}{c|}{0.220} & 0.217 & \multicolumn{1}{c|}{0.296} & 0.196           & 0.255     \\
\multicolumn{1}{c|}{}                           & \multicolumn{1}{c|}{192} & \underline {0.217}                           & \multicolumn{1}{c|}{{\color[HTML]{000000} \textbf{0.247}}} & {\color[HTML]{000000} \textbf{0.216}} & \multicolumn{1}{c|}{\underline {0.251}}                           & 0.240          & \multicolumn{1}{c|}{0.269} & 0.231                                 & \multicolumn{1}{c|}{0.263}       & 0.221                                 & \multicolumn{1}{c|}{0.254}                                 & 0.225                                 & \multicolumn{1}{c|}{0.259}                                 & 0.219       & \multicolumn{1}{c|}{0.261} & 0.276 & \multicolumn{1}{c|}{0.336} & 0.237           & 0.296     \\
\multicolumn{1}{c|}{}                           & \multicolumn{1}{c|}{336} & \underline {0.275}                           & \multicolumn{1}{c|}{{\color[HTML]{000000} \textbf{0.289}}} & {\color[HTML]{000000} \textbf{0.271}} & \multicolumn{1}{c|}{\underline {0.292}}                           & 0.293          & \multicolumn{1}{c|}{0.306} & 0.283                                 & \multicolumn{1}{c|}{0.300}       & 0.278                                 & \multicolumn{1}{c|}{0.296}                                 & 0.278                                 & \multicolumn{1}{c|}{0.297}                                 & 0.280       & \multicolumn{1}{c|}{0.306} & 0.339 & \multicolumn{1}{c|}{0.380} & 0.283           & 0.335     \\
\multicolumn{1}{c|}{\multirow{-4}{*}{Weather}}  & \multicolumn{1}{c|}{720} & {\color[HTML]{000000} \textbf{0.350}} & \multicolumn{1}{c|}{{\color[HTML]{000000} \textbf{0.339}}} & 0.355                                 & \multicolumn{1}{c|}{0.352}                                 & 0.368          & \multicolumn{1}{c|}{0.354} & 0.360                                 & \multicolumn{1}{c|}{0.350}       & 0.358                                 & \multicolumn{1}{c|}{0.349}                                 & \underline {0.354}                           & \multicolumn{1}{c|}{\underline {0.348}}                           & 0.365       & \multicolumn{1}{c|}{0.359} & 0.403 & \multicolumn{1}{c|}{0.428} & 0.345           & 0.381     \\ \midrule
\multicolumn{1}{c|}{}                           & \multicolumn{1}{c|}{96}  & 0.167                                 & \multicolumn{1}{c|}{\underline {0.246}}                           & {\color[HTML]{000000} \textbf{0.147}} & \multicolumn{1}{c|}{{\color[HTML]{000000} \textbf{0.240}}} & 0.204          & \multicolumn{1}{c|}{0.293} & 0.185                                 & \multicolumn{1}{c|}{0.272}       & \underline {0.148}                           & \multicolumn{1}{c|}{{\color[HTML]{000000} \textbf{0.240}}} & 0.195                                 & \multicolumn{1}{c|}{0.285}                                 & 0.168       & \multicolumn{1}{c|}{0.272} & 0.193 & \multicolumn{1}{c|}{0.308} & 0.197           & 0.282     \\
\multicolumn{1}{c|}{}                           & \multicolumn{1}{c|}{192} & 0.176                                 & \multicolumn{1}{c|}{\underline {0.254}}                           & \underline {0.163}                           & \multicolumn{1}{c|}{\underline {0.254}}                           & 0.207          & \multicolumn{1}{c|}{0.295} & 0.189                                 & \multicolumn{1}{c|}{0.276}       & {\color[HTML]{000000} \textbf{0.162}} & \multicolumn{1}{c|}{{\color[HTML]{000000} \textbf{0.253}}} & 0.199                                 & \multicolumn{1}{c|}{0.289}                                 & 0.184       & \multicolumn{1}{c|}{0.289} & 0.201 & \multicolumn{1}{c|}{0.315} & 0.196           & 0.285     \\
\multicolumn{1}{c|}{}                           & \multicolumn{1}{c|}{336} & \underline {0.190}                           & \multicolumn{1}{c|}{\underline {0.270}}                           & {\color[HTML]{000000} \textbf{0.178}} & \multicolumn{1}{c|}{\underline {0.270}}                           & 0.219          & \multicolumn{1}{c|}{0.308} & 0.204                                 & \multicolumn{1}{c|}{0.291}       & {\color[HTML]{000000} \textbf{0.178}} & \multicolumn{1}{c|}{{\color[HTML]{000000} \textbf{0.269}}} & 0.215                                 & \multicolumn{1}{c|}{0.305}                                 & 0.198       & \multicolumn{1}{c|}{0.300} & 0.214 & \multicolumn{1}{c|}{0.329} & 0.209           & 0.301     \\
\multicolumn{1}{c|}{\multirow{-4}{*}{ECL}}      & \multicolumn{1}{c|}{720} & 0.228                                 & \multicolumn{1}{c|}{\underline {0.303}}                           & {\color[HTML]{000000} \textbf{0.215}} & \multicolumn{1}{c|}{{\color[HTML]{000000} \textbf{0.300}}} & 0.263          & \multicolumn{1}{c|}{0.341} & 0.245                                 & \multicolumn{1}{c|}{0.324}       & 0.225                                 & \multicolumn{1}{c|}{0.317}                                 & 0.256                                 & \multicolumn{1}{c|}{0.337}                                 & \underline {0.220} & \multicolumn{1}{c|}{0.320} & 0.246 & \multicolumn{1}{c|}{0.355} & 0.245           & 0.333     \\ \midrule
\multicolumn{1}{c|}{}                           & \multicolumn{1}{c|}{96}  & 0.434                                 & \multicolumn{1}{c|}{{\color[HTML]{000000} \textbf{0.252}}} & \underline {0.416}                           & \multicolumn{1}{c|}{0.274}                                 & 0.536          & \multicolumn{1}{c|}{0.359} & 0.468                                 & \multicolumn{1}{c|}{0.307}       & {\color[HTML]{000000} \textbf{0.395}} & \multicolumn{1}{c|}{\underline {0.268}}                           & 0.544                                 & \multicolumn{1}{c|}{0.359}                                 & 0.593       & \multicolumn{1}{c|}{0.321} & 0.587 & \multicolumn{1}{c|}{0.366} & 0.650           & 0.396     \\
\multicolumn{1}{c|}{}                           & \multicolumn{1}{c|}{192} & 0.450                                 & \multicolumn{1}{c|}{{\color[HTML]{000000} \textbf{0.257}}} & \underline {0.430}                           & \multicolumn{1}{c|}{\underline {0.276}}                           & 0.530          & \multicolumn{1}{c|}{0.354} & 0.476                                 & \multicolumn{1}{c|}{0.311}       & {\color[HTML]{000000} \textbf{0.417}} & \multicolumn{1}{c|}{\underline {0.276}}                           & 0.540                                 & \multicolumn{1}{c|}{0.354}                                 & 0.617       & \multicolumn{1}{c|}{0.336} & 0.604 & \multicolumn{1}{c|}{0.373} & 0.598           & 0.370     \\
\multicolumn{1}{c|}{}                           & \multicolumn{1}{c|}{336} & 0.466                                 & \multicolumn{1}{c|}{{\color[HTML]{000000} \textbf{0.264}}} & \underline {0.451}                           & \multicolumn{1}{c|}{0.286}                                 & 0.530          & \multicolumn{1}{c|}{0.349} & 0.488                                 & \multicolumn{1}{c|}{0.317}       & {\color[HTML]{000000} \textbf{0.433}} & \multicolumn{1}{c|}{\underline {0.283}}                           & 0.551                                 & \multicolumn{1}{c|}{0.358}                                 & 0.629       & \multicolumn{1}{c|}{0.336} & 0.621 & \multicolumn{1}{c|}{0.383} & 0.605           & 0.373     \\
\multicolumn{1}{c|}{\multirow{-4}{*}{Traffic}}  & \multicolumn{1}{c|}{720} & 0.500                                 & \multicolumn{1}{c|}{{\color[HTML]{000000} \textbf{0.283}}} & \underline {0.478}                           & \multicolumn{1}{c|}{\underline {0.301}}                           & 0.569          & \multicolumn{1}{c|}{0.371} & 0.521                                 & \multicolumn{1}{c|}{0.333}       & {\color[HTML]{000000} \textbf{0.467}} & \multicolumn{1}{c|}{0.302}                                 & 0.586                                 & \multicolumn{1}{c|}{0.375}                                 & 0.640       & \multicolumn{1}{c|}{0.350} & 0.626 & \multicolumn{1}{c|}{0.382} & 0.645           & 0.394     \\ \midrule
\multicolumn{1}{c|}{}                           & \multicolumn{1}{c|}{96}  & {\color[HTML]{000000} \textbf{0.081}} & \multicolumn{1}{c|}{{\color[HTML]{000000} \textbf{0.198}}} & 0.083                                 & \multicolumn{1}{c|}{0.203}                                 & 0.123          & \multicolumn{1}{c|}{0.251} & 0.096                                 & \multicolumn{1}{c|}{0.218}       & 0.086                                 & \multicolumn{1}{c|}{0.206}                                 & 0.088                                 & \multicolumn{1}{c|}{0.205}                                 & 0.107       & \multicolumn{1}{c|}{0.234} & 0.148 & \multicolumn{1}{c|}{0.278} & 0.088           & 0.218     \\
\multicolumn{1}{c|}{}                           & \multicolumn{1}{c|}{192} & {\color[HTML]{000000} \textbf{0.176}} & \multicolumn{1}{c|}{{\color[HTML]{000000} \textbf{0.295}}} & 0.186                                 & \multicolumn{1}{c|}{0.306}                                 & 0.224          & \multicolumn{1}{c|}{0.344} & 0.182                                 & \multicolumn{1}{c|}{0.307}       & 0.177                                 & \multicolumn{1}{c|}{0.299}                                 & {\color[HTML]{000000} \textbf{0.176}} & \multicolumn{1}{c|}{0.299}                                 & 0.226       & \multicolumn{1}{c|}{0.344} & 0.271 & \multicolumn{1}{c|}{0.315} & 0.176           & 0.315     \\
\multicolumn{1}{c|}{}                           & \multicolumn{1}{c|}{336} & 0.347                                 & \multicolumn{1}{c|} {\underline {0.424}}                          & 0.350                                 & \multicolumn{1}{c|}{0.427}                                 & 0.377          & \multicolumn{1}{c|}{0.451} & 0.402                                 & \multicolumn{1}{c|}{0.461}       & 0.331                                 & \multicolumn{1}{c|}{{\color[HTML]{000000} \textbf{0.397}}} & {\color[HTML]{000000} \textbf{0.301}} & \multicolumn{1}{c|}{{\color[HTML]{000000} \textbf{0.397}}} & 0.367       & \multicolumn{1}{c|}{0.448} & 0.460 & \multicolumn{1}{c|}{0.427} & \underline {0.313}     & 0.427     \\
\multicolumn{1}{c|}{\multirow{-4}{*}{Exchange}} & 720                      & {\color[HTML]{000000} \textbf{0.795}} & {\color[HTML]{000000} \textbf{0.668}}                      & 0.935                                 & 0.732                                                      & 1.018          & 0.771                      & 1.055                                 & 0.767                            & 0.847                                 & 0.691                                                      & 0.901                                 & 0.714                                                      & 0.964       & 0.746                      & 1.195 & 0.695                      & 0.839           & 0.695     \\ \bottomrule
\end{tabular}
\end{table}

\noindent \textbf{Long-term Forecasting.}\hspace{0.2cm}We conduct evaluations on 9 benchmark datasets to assess long-term forecasting. As shown in Table \ref{tab:long-term}, Logo-LLM achieves better performance than most baselines. Specifically, compared to the recent SOTA methods CALF and Time-LLM, Logo-LLM yields a relative MSE reduction of 1.3\% and 8.9\%, respectively.

\noindent \textbf{Few-shot Forecasting.}\hspace{0.2cm}With extensive world knowledge, LLMs exhibit excellent few-shot and zero-shot learning abilities. In NLP, prompting with a few examples can often yield comparable performance with fine-tuning. This motivates the exploration of LLMs' generalization capabilities in time series forecasting under limited supervision. Following the setup of ~\cite{zhou2023onefitsall}, each dataset is divided into training, validation, and test sets, with only 5\% training data used in few-shot scenarios. We evaluate Logo-LLM primarily on four ETT benchmarks to validate its effectiveness.

\begin{table}[pos=htbp]\scriptsize
\centering
\caption{Few-shot learning task on 5\% data. The results are averaged from prediction lengths $T \in \{96, 192, 336, 720\}$.}
\label{tab:few-shot5}
\renewcommand{\arraystretch}{1}
\setlength{\tabcolsep}{3.5pt}
\linespread{1}
\begin{tabular}{@{}c|cc|cc|cc|cc|cc|cc|cc|cc|cc@{}}
\toprule
Methods & \multicolumn{2}{c|}{\makecell{Logo-LLM\\ (Ours)}}                                                 & \multicolumn{2}{c|}{CALF} & \multicolumn{2}{c|}{Time-LLM} & \multicolumn{2}{c|}{GPT4TS} & \multicolumn{2}{c|}{iTransformer} & \multicolumn{2}{c|}{PatchTST}                                                 & \multicolumn{2}{c|}{TimesNet} & \multicolumn{2}{c|}{FEDformer} & \multicolumn{2}{c}{DLinear} \\ \midrule
Metric  & MSE                                   & MAE                                   & MSE            & MAE            & MSE           & MAE          & MSE     & MAE              & MSE             & MAE            & MSE                                   & MAE                                   & MSE           & MAE          & MSE                & MAE      & MSE      & MAE              \\ \midrule
ETTm1   & 0.540                        & {\color[HTML]{000000} \textbf{0.473}} & 0.611          & 0.512          & 0.648         & 0.527        & 0.626   & 0.510            & 0.731           & 0.567          & {\color[HTML]{000000} \textbf{0.526}} & 0.476 & 0.717         & 0.561        & 0.730              & 0.592    & 0.572    & 0.502   \\
ETTm2   & {\color[HTML]{000000} \textbf{0.306}} & {\color[HTML]{000000} \textbf{0.339}} & 0.314 & 0.350 & 0.318         & 0.353        & 0.328   & 0.357            & 0.343           & 0.369          & 0.314                        & 0.352                                 & 0.344         & 0.372        & 0.381              & 0.404    & 0.510    & 0.492            \\
ETTh1   & {\color[HTML]{000000} \textbf{0.606}} & {\color[HTML]{000000} \textbf{0.519}} & 0.840          & 0.613          & 0.863         & 0.640        & 0.673   & 0.557   & 0.831           & 0.624          & 0.694                                 & 0.569                                 & 0.925         & 0.647        & 0.658     & 0.562    & 0.750    & 0.611            \\
ETTh2   & {\color[HTML]{000000} \textbf{0.394}} & {\color[HTML]{000000} \textbf{0.407}} & 0.457          & 0.447 & 0.548         & 0.478        & 0.523   & 0.473            & 0.472           & 0.458          & 0.439                        & 0.448                                 & 0.463         & 0.454        & 0.441              & 0.457    & 0.827    & 0.615            \\ \bottomrule
\end{tabular}
\end{table}

The results of 5\% few-shot learning with input length $L=96$ are shown in Table \ref{tab:few-shot5}. Compared with LLM-based methods (~\cite{liu2025calf,jin2023timellm,zhou2023onefitsall}), our proposed Logo-LLM achieves the best or second best prediction performance on all datasets. This demonstrates its excellent ability to effectively capture both local and global patterns, even under low-data settings. In comparison to CALF (~\cite{liu2025calf}) and Time-LLM (~\cite{jin2023timellm}), Logo-LLM demonstrates a relative average MSE reduction of 13.5\% and 19.2\% respectively. These results highlight the strong data efficiency and robust generalization of Logo-LLM in low-resource forecasting scenarios.

\noindent \textbf{Zero-shot Forecasting.}\hspace{0.2cm}
\begin{table}[pos=H]\scriptsize
\centering
\caption{Zero-shot learning task. The results are averaged from prediction lengths $T \in \{96, 192, 336, 720\}$. A $\rightarrow$ B indicates a zero-shot transfer setting where the model is trained on dataset A and directly evaluated on dataset B without further fine-tuning.}
\label{tab:zero-shot}
\renewcommand{\arraystretch}{1}
\setlength{\tabcolsep}{1.8pt}
\linespread{1}
\begin{tabular}{@{}c|cc|cc|cc|cc|cc|cc|cc|cc|cc@{}}
\toprule
Methods & \multicolumn{2}{c|}{\makecell{Logo-LLM\\ (Ours)}}                                                 & \multicolumn{2}{c|}{\makecell{CALF}} & \multicolumn{2}{c|}{\makecell{Time-LLM}} & \multicolumn{2}{c|}{\makecell{GPT4TS}} & \multicolumn{2}{c|}{\makecell{iTransformer}} & \multicolumn{2}{c|}{\makecell{PatchTST}}                                                 & \multicolumn{2}{c|}{\makecell{TimesNet}} & \multicolumn{2}{c|}{\makecell{FEDformer}} & \multicolumn{2}{c}{\makecell{DLinear}} \\ \midrule
Metric  & MSE                                    & MAE                                   & MSE            & MAE            & MSE           & MAE           & MSE            & MAE            & MSE             & MAE             & MSE                                   & MAE            & MSE           & MAE           & MSE            & MAE           & MSE          & MAE          \\ \midrule
ETTh1 $\rightarrow$ ETTm1 & 0.762                                 & {\color[HTML]{000000} \textbf{0.566}} & {\color[HTML]{000000} \textbf{0.755}} & 0.574 & 0.847         & 0.565         & 0.798        & 0.574        &    0.825             &      0.589           & 0.894         & 0.610         & 0.794         & 0.575         & 0.765                                 & 0.588 & 0.760    & 0.577   \\
ETTh1 $\rightarrow$ ETTm2 & {\color[HTML]{000000} \textbf{0.313}} & {\color[HTML]{000000} \textbf{0.353}} & 0.316                    & 0.355 & 0.315         & 0.357         & 0.317        & 0.359        &      0.320           &     0.359            & 0.318         & 0.362         & 0.339         & 0.370         & 0.357                                 & 0.403 & 0.399             & 0.439   \\
ETTh2 $\rightarrow$ ETTm1 & 0.841                                 & {\color[HTML]{000000} \textbf{0.578}} & 0.836                                 & 0.586 & 0.868         & 0.595         & 0.920        & 0.610        &     0.912            &       0.603          & 0.871         & 0.596         & 1.286         & 0.705         & {\color[HTML]{000000} \textbf{0.741}} & 0.588 & 0.778    & 0.594   \\
ETTh2 $\rightarrow$ ETTm2 & {\color[HTML]{000000} \textbf{0.316}} & {\color[HTML]{000000} \textbf{0.357}} & 0.319                                 & 0.360 & 0.322         & 0.363         & 0.331        & 0.371        &     0.329            &   0.370              & 0.420         & 0.433         & 0.361         & 0.390         & 0.365                                 & 0.405 & 0.496             & 0.496   \\ \bottomrule
\end{tabular}
\end{table}
We further investigate the zero-shot generalization ability of LLMs from ~\cite{liu2025calf}, where models are evaluated on a completely unseen dataset after being trained on a different source dataset, without any fine-tuning. This setup highlights the model's capacity for domain transfer and reasoning without task-specific supervision. As shown in Table \ref{tab:zero-shot}, Logo-LLM achieves superior performance in almost all scenarios. These results demonstrate the cross-domain generalization capability of our approach, underscoring its effectiveness in zero-shot learning.

\section{Ablation Studies}
We have conducted several ablation studies to analyze Logo-LLM. Specially, the experiments cover model fine-tuning, parameter configuration, feature mixer strategy, input length, different LLMs, and the strategy of local and global modeling.

\begin{table}[pos=htbp]\scriptsize
\centering
\caption{Ablations of 5\% training data on fine-tune strategy. We select several methods to fine-tune the pre-trained LLM. The average results for all prediction lengths are listed here.}
\label{tab:fine-tune}
\renewcommand{\arraystretch}{1}
\setlength{\tabcolsep}{1.5pt}
\linespread{1}
\begin{tabular}{@{}c|cccccccccccc@{}}
\toprule
Methods & \multicolumn{2}{c}{LN+PE} & \multicolumn{2}{c}{LN} & \multicolumn{2}{c}{PE} & \multicolumn{2}{c}{FP} & \multicolumn{2}{c}{LoRA} & \multicolumn{2}{c}{w/o FT} \\ \midrule
Metric  & MSE           & MAE          & MSE        & MAE       & MSE        & MAE       & MSE           & MAE           & MSE         & MAE        & MSE             & MAE             \\ \midrule
ETTm1   & \textbf{0.540}         & 0.473        & 0.543      & 0.474     & 0.541      & 0.474     & 0.537         & \textbf{0.472}         & 0.549       & 0.478      & 0.541           & 0.474           \\
ETTm2   & \textbf{0.306}         & \textbf{0.339}        & 0.308      & 0.341     & 0.308      & 0.341     & 0.330         & 0.351         & 0.308       & 0.341      & 0.308           & 0.341           \\
ETTh1   & \textbf{0.606}         & \textbf{0.519}        & 0.640      & 0.542     & 0.640      & 0.542     & 0.642         & 0.543         & 0.640       & 0.543      & 0.640           & 0.542           \\
ETTh2   & \textbf{0.394}         & \textbf{0.407}        & 0.399      & \textbf{0.407}     & 0.399      & \textbf{0.407}     & 0.403         & 0.410         & 0.398       & \textbf{0.407}      & 0.399           & \textbf{0.407}           \\ \bottomrule
\end{tabular}
\end{table}

\noindent \textbf{Fine-tuning Strategy.}\hspace{0.2cm}To assess different fine-tuning strategies, we conducted ablation experiments using only 5\% of training data across ETT datasets. As shown in Table \ref{tab:fine-tune}, we compare full-parameter (FP), LoRA (~\cite{hu2022lora}), frozen fine-tuning of layer normalization and positional embedding layer (LN+PE), and the baseline, without fine-tuning (w/o FT). Among the strategies, LN+PE consistently achieves superior or comparable performance across all datasets and metrics. In contrast, FP fails to show a clear advantage and even slightly underperforms on ETTm2 and ETTh2, possibly due to overfitting in small-scale datasets.

\noindent \textbf{Feature Mixer Strategy.}\hspace{0.2cm}Table \ref{tab:mixer} presents ablation results on 5\% training data to evaluate different strategies for aligning local and global features with temporal data. Specifically, we compare our proposed Mixer modules against addition (Add), cross-attention mechanisms (Cross), and a baseline without any fusion (w/o Mixer). The results show that Mixer can consistently achieve superior performance across all datasets after aligning the local and global variations, respectively, highlighting its effectiveness in capturing fine-grained interactions between temporal inputs and hierarchical feature representations from the pre-trained LLM.

\begin{table}[pos=htbp]\scriptsize
\centering
\caption{Ablations of 5\% training data on local and global features. We select different methods to align the temporal data with local and global features, respectively. The average results of all prediction lengths are listed here.}
\label{tab:mixer}
\renewcommand{\arraystretch}{1}
\setlength{\tabcolsep}{5pt}
\linespread{1}
\begin{tabular}{@{}c|cccccccc@{}}
\toprule
Methods & \multicolumn{2}{c}{Mixer (Ours)}                                   & \multicolumn{2}{c}{Add} & \multicolumn{2}{c}{Cross} & \multicolumn{2}{c}{w/o Mixer} \\ \midrule
Metric  & MSE                          & MAE                          & MSE        & MAE        & MSE              & MAE              & MSE           & MAE           \\ \midrule
ETTm1   & \textbf{0.540} & \textbf{0.473} & 0.557      & \textbf{0.473}      & 0.555            & 0.492            & 0.593         & 0.500         \\
ETTm2   & \textbf{0.306}         & \textbf{0.339} & 0.333      & 0.360      & 0.314            & 0.352            & 0.316         & 0.350         \\
ETTh1   & \textbf{0.606}         & \textbf{0.519} & 0.685      & 0.560      & 0.652            & 0.551            & 0.648         & 0.551         \\
ETTh2   & \textbf{0.394}         & \textbf{0.407} & 0.564      & 0.484      & 0.496            & 0.451            & 0.593         & 0.495         \\ \bottomrule
\end{tabular}
\end{table}

\noindent \textbf{The Number of LLM Layers.}\hspace{0.2cm}To leverage the representational ability of the pre-trained LLM while controlling computational overhead, we select a subset of layers to balance performance and efficiency. Therefore, we conduct ablations on numbers of LLM layers on ETTm1 and ETTm2 using 5\% training data. Additionally, we choose the LLM-based method CALF (~\cite{liu2025calf}) as a comparison.

\begin{figure}[!htbp]
  \centering
  \includegraphics[width=1\linewidth]{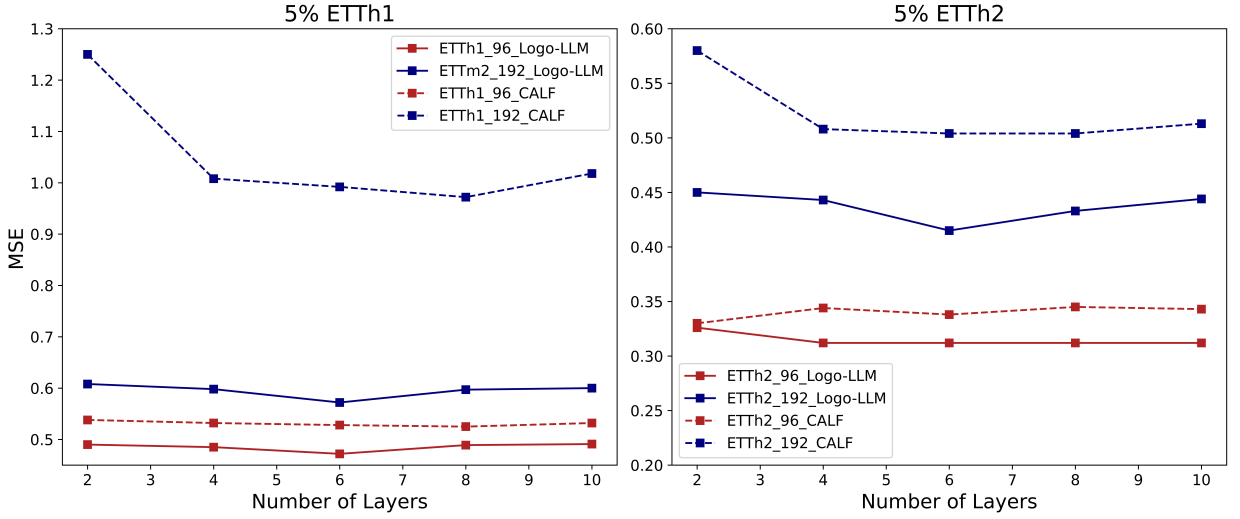}
  \caption{Comparison of Logo-LLM and CALF with various layers on ETTh1 and ETTh2 datasets. The prediction length is set as \{96, 192\}.}
  \label{fig:layer_visual}
  \vspace{-0.2cm}
\end{figure}

As shown in Figure \ref{fig:layer_visual}, just two Transformer layers are sufficient to unlock most of the LLM's representational power for time series modeling. As the depth increases, the predictive performance improves and reaches its peak around 6 layers. This highlights that early layers in the pre-trained LLM already encode rich local and global temporal patterns. Moreover, there exists a sweet spot in model depth where the trade-off between long-term dependencies and fine-grained variations is balanced. This observation validates our design of selectively leveraging shallow and deep layer representations, rather than relying on the last layer.

\begin{figure}[!htbp]
  \centering
  \includegraphics[width=1\linewidth]{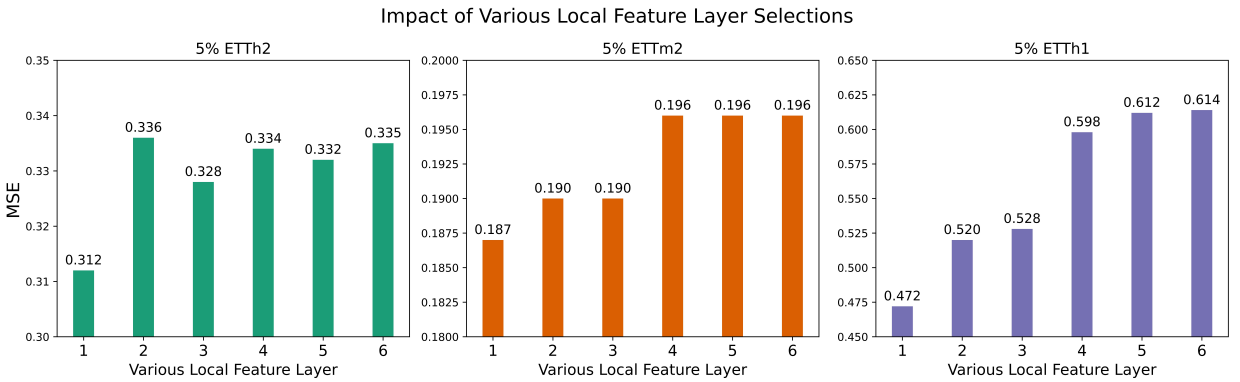}
  \caption{Visualization of different selections \{1, 2, 3, 4, 5, 6\} about local feature layer on ETTh1, ETTm2, and ETTh2. The prediction length is set as 96 with input length $L=96$.}
  \label{fig:local_visual}
\end{figure}

\noindent \textbf{Impact of Local Feature Layer Selection.}\hspace{0.2cm}To investigate the optimal layer for extracting local representations from the pre-trained LLM, we conducted an ablation study by varying the selected Transformer layer to represent local features. As shown in Figure \ref{fig:local_visual}, we evaluated the performance of Logo-LLM with MSE as the evaluation criterion.

\begin{figure}[!htbp]
  \centering
  \subfloat[Logo-LLM\label{fig:logo_at}]{
    \includegraphics[width=0.28\linewidth]{logo21.pdf}
  }
  \hfill
  \subfloat[CALF\label{fig:calf_at}]{
    \includegraphics[width=0.28\linewidth]{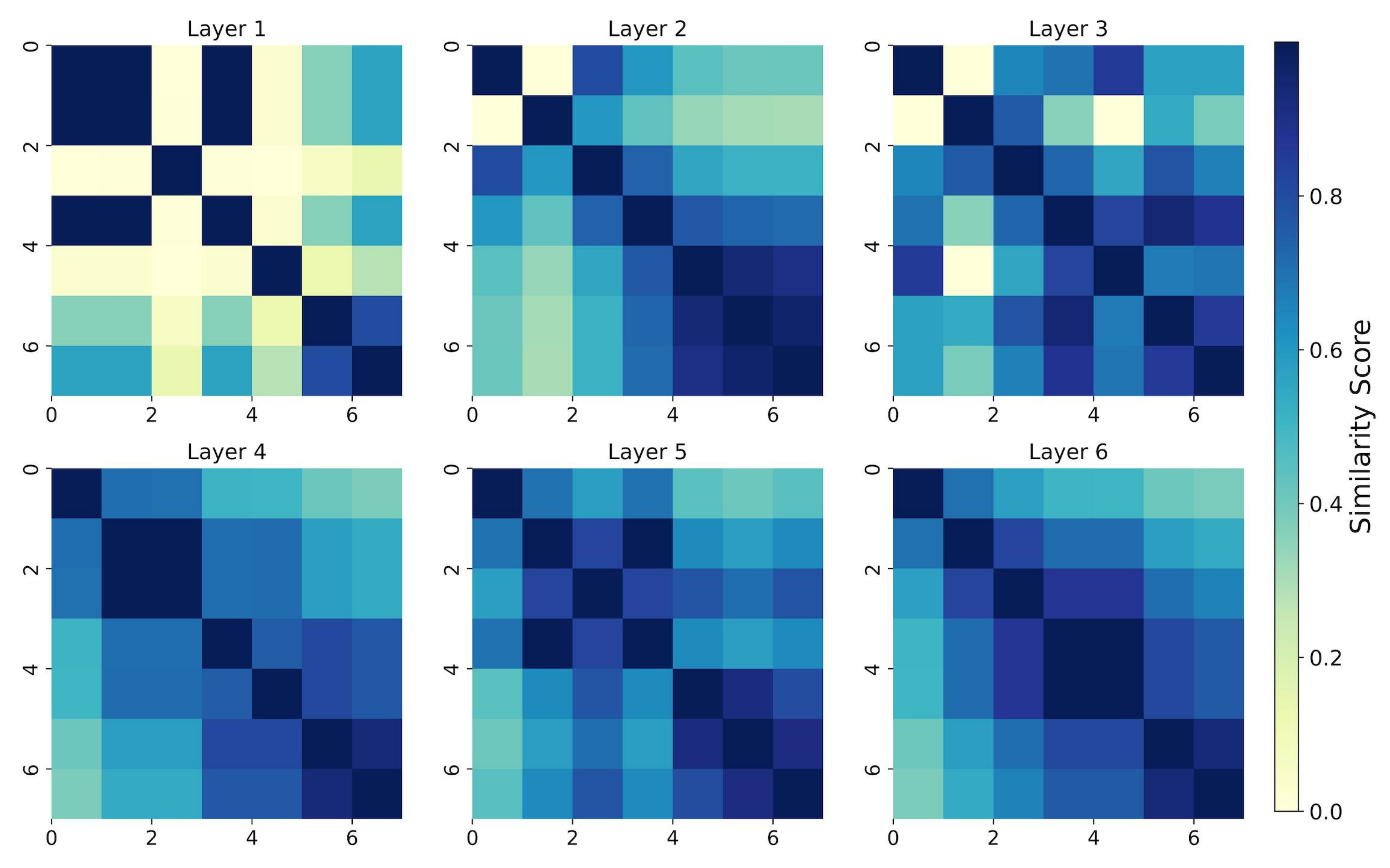}
  }
  \hfill
  \subfloat[Time-LLM\label{fig:time_at}]{
    \includegraphics[width=0.28\linewidth]{timellm.pdf}
  }
  \caption{Similarity matrices of each patch across Transformer layers in (a) Logo-LLM (b) CALF and (c) Time-LLM, illustrating that shallow layers exhibit pronounced local patterns while deeper layers capture broader global dependencies.}
  \label{fig:logo-att}
\end{figure}

We observe that using the first-layer output as a local feature yields the best performance and performance gradually deteriorates or plateaus when deeper layers are used. This finding supports our design choice and aligns with the representational hierarchy of LLMs. Early Transformer layer in LLMs tend to preserve fine-grained temporal patterns that are more directly correlated with the original input, whereas deeper layers abstract away local variations in favor of global semantics. By designing local modeling at the shallowest layer, Logo-LLM can effectively capture short-term fluctuations that are critical for time series forecasting.

\noindent \textbf{Ablation on Large Language Models.}\hspace{0.2cm}To further validate our findings, we extended our experiments to apply LLMs with diverse architectures, i.e., the encoder-based BERT (\cite{devlin2019bert}). As shown in Table \ref{tab:llm}, the performance gap between the two models across multiple ETT datasets is negligible. This indicates that hierarchical specialization capabilities, where shallow layers capture local patterns and deeper layers encode global dependencies, are not unique to GPT-2 (\cite{radford2019gpt2}). Instead, this capability exists as a universal intrinsic property of LLMs, independent of specific architectural designs.

\begin{table}[pos=htbp]\scriptsize
\centering
\caption{Ablations on different LLMs with the input length $L = 96$.}
\label{tab:llm}
\renewcommand{\arraystretch}{1}
\linespread{1}
\begin{tabular}{@{}cc|cc|cc@{}}
\toprule
\multicolumn{2}{c|}{Methods}                      & \multicolumn{2}{c|}{GPT-2} & \multicolumn{2}{c}{BERT} \\ \midrule
\multicolumn{2}{c|}{Metric}                       & MSE           & MAE           & MSE                                   & MAE                                   \\ \midrule
\multicolumn{1}{c|}{\multirow{4}{*}{ETTm1}} & 96  & 0.317         & 0.343         & 0.316                                 & 0.343                                 \\
\multicolumn{1}{c|}{}                       & 192 & 0.368         & 0.370         & 0.368                                 & 0.372                                 \\
\multicolumn{1}{c|}{}                       & 336 & 0.394         & 0.393         & 0.392                                 & 0.394                                 \\
\multicolumn{1}{c|}{}                       & 720 & 0.460         & 0.430         & 0.453                                 & 0.430                                 \\ \midrule
\multicolumn{1}{c|}{\multirow{4}{*}{ETTm2}} & 96  & 0.175         & 0.253         & 0.174                                 & 0.251                                 \\
\multicolumn{1}{c|}{}                       & 192 & 0.243         & 0.298         & 0.241                                 & 0.296                                 \\
\multicolumn{1}{c|}{}                       & 336 & 0.304         & 0.338         & 0.302                                 & 0.336                                 \\
\multicolumn{1}{c|}{}                       & 720 & 0.408         & 0.400         & 0.402                                 & 0.394                                 \\ \midrule
\multicolumn{1}{c|}{\multirow{4}{*}{ETTh1}} & 96  & 0.373         & 0.388         & 0.372                                 & 0.391                                 \\
\multicolumn{1}{c|}{}                       & 192 & 0.413         & 0.412         & 0.419                                 & 0.418                                 \\
\multicolumn{1}{c|}{}                       & 336 & 0.434         & 0.424         & 0.454                                 & 0.439                                 \\
\multicolumn{1}{c|}{}                       & 720 & 0.446         & 0.447         & 0.457                                 & 0.456                                 \\ \midrule
\multicolumn{1}{c|}{\multirow{4}{*}{ETTh2}} & 96  & 0.274         & 0.330         & 0.298                                 & 0.347                                 \\
\multicolumn{1}{c|}{}                       & 192 & 0.344         & 0.376         & 0.361                                 & 0.387                                 \\
\multicolumn{1}{c|}{}                       & 336 & 0.369         & 0.398         & 0.374                                 & 0.399                                 \\
\multicolumn{1}{c|}{}                       & 720 & 0.394         & 0.417         & 0.392                                 & 0.423                                 \\ \bottomrule
\end{tabular}
\end{table}

\noindent \textbf{Local and Global Modeling.}\hspace{0.2cm}Table \ref{tab:logo} presents the ablation results using 5\% training data to analyze local and global feature modeling strategies. The default Logo-LLM leverages the first and last LLM layers for local and global representations, respectively. We compare this with a variant (denoted by $^*$) that uses the average of the first and second half from all LLM layers as local and global features, respectively. The results demonstrate that selecting the first and last layers yields better performance, indicating that boundary-layer features capture more distinct local and global variations than uniformly averaged intermediate representations.

\begin{table}[pos=htbp]\scriptsize
\centering
\caption{Ablations on modeling strategy. w/o represents the removal of the corresponding representation, and $^*$ denotes using the averaged representations from first half of LLM layers as local features, and those from second half as global features.}
\label{tab:logo}
\renewcommand{\arraystretch}{1}
\setlength{\tabcolsep}{5pt}
\linespread{1}
\begin{tabular}{@{}c|cccccccccccc@{}}
\toprule
Methods & \multicolumn{2}{c}{Logo-LLM} & \multicolumn{2}{c}{w/o Global} & \multicolumn{2}{c}{w/o Local} & \multicolumn{2}{c}{Logo-LLM$^*$} & \multicolumn{2}{c}{w/o Global$^*$} & \multicolumn{2}{c}{w/o Local$^*$} \\ \midrule
Metric  & MSE           & MAE          & MSE            & MAE           & MSE           & MAE           & MSE       & MAE       & MSE            & MAE            & MSE            & MAE           \\ \midrule
ETTm1   & 0.540         & 0.473        & 0.586          & 0.486         & 0.611         & 0.497         & 0.573     & 0.478     & \textbf{0.519}          & \textbf{0.459}          & 0.552          & 0.467         \\
ETTm2   & \textbf{0.306}         & \textbf{0.339}        & 0.343          & 0.364         & 0.335         & 0.360         & 0.310     & 0.344     & 0.330          & 0.354          & 0.311          & 0.343         \\
ETTh1   & \textbf{0.606}         & \textbf{0.519}        & 0.642          & 0.545         & 0.644         & 0.549         & 0.635     & 0.539     & 0.707          & 0.575          & 0.959          & 0.649         \\
ETTh2   & \textbf{0.394}         & \textbf{0.407}        & 0.414          & 0.414         & 0.403         & 0.414         & 0.586     & 0.493     & 0.612          & 0.619          & 0.604          & 0.499         \\ \bottomrule
\end{tabular}
\end{table}

\noindent \textbf{Local and Global Representation Analysis.}\hspace{0.2cm}To provide empirical validation for our hypothesis that shallow and deep layers of LLM specialize in the extraction of local and global patterns, we visualize the cosine similarity matrices of hidden representations across all six layers in Figure \ref{fig:logo-att}. In the early layers (Layers 1--2), we observe that similarity scores are highly concentrated around the diagonal. This indicates that each patch predominantly attends to its temporal neighbors, confirming that shallow layers focus on short-term patterns and local consistency. In contrast, some rows in the similarity matrix exhibit local clustering patterns, where high similarity scores appear in off-diagonal regions. This behavior is reminiscent of how LLMs process textual data, capturing syntactic or surface-level relationships in earlier layers and progressively learning semantic or abstract representations in deeper layers (~\cite{lee2024causal, liu2024fantastic}). This suggests that the model has captured local structures, even when the corresponding patches are not temporally adjacent in the sequence.

As the model progresses to deeper layers (Layers 3--6), the similarity matrices exhibit increasingly smooth and uniformly distributed patterns. Elevated similarity is observed even between distant patches, indicating a transition from localized representations toward more global and abstract ones. At this stage, the model begins to integrate information across the entire sequence, allowing it to capture long-range dependencies, overarching trends structures. This layer-wise transition, from local pattern extraction in the shallow layers to global temporal abstraction in the deeper layers, is well aligned with the architectural design of Logo-LLM. In particular, we employ a Local Mixer on early-layer outputs to effectively capture short-term local variations, and a Global Mixer on the deeper-layer representations to model long-range dependencies. This hierarchical decomposition enables Logo-LLM to learn temporal dynamics at multiple scales, enhancing its ability to respond to short-term fluctuations while maintaining long-range forecasting capabilities.

\begin{table}[pos=htbp]\scriptsize
\centering
\caption{Comparison of LLM-based time series
forecasting methods in terms of total parameters (Param.), training parameters ($\text{Param.}^*$), computation time for a single run and MSE. The input and prediction lengths are set as 96 and 720 respectively.}
\label{tab:time}
\renewcommand{\arraystretch}{1.5}
\setlength{\tabcolsep}{1.6pt}
\linespread{1}
\begin{tabular}{@{}c|cccc|cccc|cccc|cccc@{}}
\toprule
\multirow{2}{*}{Methods} & \multicolumn{4}{c|}{\makecell{Logo-LLM\\ (Ours)}}              & \multicolumn{4}{c|}{\makecell{CALF}}                    & \multicolumn{4}{c|}{\makecell{Time-LLM}}               & \multicolumn{4}{c}{\makecell{GPT4TS}}                 \\ \cmidrule(l){2-17} 
                         & Param. & $\text{Param.}^*$ & Time  & MSE     & Param.  & $\text{Param.}^*$ & Time   & MSE     & Param.   & $\text{Param.}^*$ & Time & MSE     & Param. & $\text{Param.}^*$ & Time  & MSE     \\ \midrule
Weather                  & 92.1M & 11.0M       & 1.05ms & 0.350 & 180.7M & 18.5M       & 10.95ms & 0.355 & 6653.1M & 45.7M       &   192.3ms    & 0.368 & 88.6M & 7.5M        & 0.93ms & 0.360 \\
ETTh1                    & 92.1M & 11.0M       & 0.83ms & 0.446 & 180.7M & 18.5M       & 8.99ms  & 0.479 & 6653.1M & 45.7M       &   210.6ms    & 0.483 & 88.6M & 7.5M        & 0.76ms & 0.495 \\ \bottomrule
\end{tabular}
\end{table}

\noindent \textbf{Time Cost Analysis.}\hspace{0.2cm}We conducted experiments about time cost on two datasets: Weather and ETTh1, and the input and prediction lengths are set to 96 and 720, respectively. The batch size is set to 1 and we test all models on same GPU. As shown in Table \ref{tab:time}, Logo-LLM shows significant improvements in both efficiency and accuracy compared to CALF (~\cite{liu2025calf}) and Time-LLM (~\cite{jin2023timellm}). Moreover, Logo-LLM performs more fine-grained modeling of local and global temporal features than GPT4TS (~\cite{zhou2023onefitsall}). Although this leads to a marginal increase in inference latency, it results in a greater improvement in predictive performance.

\section{Conclusion}
In this paper, we propose Logo-LLM, a novel framework that utilizes the hierarchical feature extraction of pre-trained large language models (LLMs) for time series forecasting. Unlike previous approaches, Logo-LLM leverages both shallow and deep layers of LLMs to capture local and global temporal features separately. The Local-Mixer and Global-Mixer modules align these features with time series data, enhancing the model's ability to capture fine-grained fluctuations and long-range dependencies. The experiments across real-world datasets demonstrate that Logo-LLM outperforms existing methods in long-term, few-shot, and zero-shot learning tasks, offering superior generalization and low computational overhead. This work highlights the potential of LLMs in time series modeling and opens avenues for further improvements in forecasting accuracy and efficiency.







\printcredits

\bibliographystyle{cas-model2-names}

\bibliography{cas-refs}
\nocite{*}



\end{document}